\documentclass[10pt,twocolumn,letterpaper]{article}

\usepackage[pagenumbers]{wacv} 

\usepackage[accsupp]{axessibility} 
\usepackage{graphicx}
\usepackage{amsmath,bm}
\usepackage{amssymb}
\usepackage{booktabs}
\usepackage{tikz,pgfplots}
\usepackage{color, colortbl}
\usepackage{float,array,multirow}
	
\definecolor{Gray}{gray}{0.9}
\usepackage{arydshln}
\usepackage[pagebackref,breaklinks,colorlinks]{hyperref}

\usepackage[capitalize]{cleveref}
\crefname{section}{Sec.}{Secs.}
\Crefname{section}{Section}{Sections}
\Crefname{table}{Table}{Tables}
\crefname{table}{Tab.}{Tabs.}

\def\wacvPaperID{1735} 
\def\confName{WACV}
\def\confYear{2025}

\begin{document}

\title{Utilizing Uncertainty in 2D Pose Detectors for Probabilistic \\3D Human Mesh Recovery}

\author{Tom Wehrbein\textsuperscript{1}
\and
Marco Rudolph\textsuperscript{1}
\and
Bodo Rosenhahn\textsuperscript{1}
\and
Bastian Wandt\textsuperscript{2}
\and
\textsuperscript{1}Leibniz University Hannover, \textsuperscript{2}Linköping University\\
{\tt\small wehrbein@tnt.uni-hannover.de}
}
\maketitle

\begin{abstract}
Monocular 3D human pose and shape estimation is an inherently ill-posed problem due to depth ambiguities, occlusions, and truncations.
Recent probabilistic approaches learn a distribution over plausible 3D human meshes by maximizing the likelihood of the ground-truth pose given an image.
We show that this objective function alone is not sufficient to best capture the full distributions.
Instead, we propose to additionally supervise the learned distributions by minimizing the distance to distributions encoded in heatmaps of a 2D pose detector.
Moreover, we reveal that current methods often generate incorrect hypotheses for invisible joints which is not detected by the evaluation protocols.
We demonstrate that person segmentation masks can be utilized during training to significantly decrease the number of invalid samples and introduce two metrics to evaluate it.
Our normalizing flow-based approach predicts plausible 3D human mesh hypotheses that are consistent with the image evidence while maintaining high diversity for ambiguous body parts.
Experiments on 3DPW and EMDB show that we outperform other state-of-the-art probabilistic methods.
Code is available for research purposes at \href{https://github.com/twehrbein/humr}{https://github.com/twehrbein/humr}.

\end{abstract}

\section{Introduction}
\label{sec:intro}
Reconstructing 3D human pose and shape from monocular images is a long-standing computer vision problem with vast applications in \eg robotics, medicine, sports, AR/VR and animation.
It is a fundamentally ill-posed problem, because multiple 3D bodies can explain a given 2D image.
Apart from the inherent depth ambiguity, body parts are often occluded or truncated and thus prevent the existence of a single unique solution.
Consequently, there has been an increasing interest in either estimating a fixed-size set of solutions \cite{biggs2020multibodies,jahangiri2017iccvw,li2019mdn,oikarinen2020graphmdn,li2022mhformer} or, more recently, modeling the full 3D pose distribution conditioned on the 2D input \cite{kolotouros2021prohmr,wehrbein2021iccv,sengupta2023humaniflow,holmquist2023diffpose,sharma2019iccv,sengupta2021prob,sengupta2021hierprob,xu2024scorehypo,chen2023mhentropy}.
In addition to the theoretical advantages, several downstream tasks like autonomous driving and video surveillance can benefit from having access to multiple plausible solutions and their prediction uncertainties. 
Furthermore, methods estimating a distribution over body parameters can act as an effective conditional prior for parametric model fitting~\cite{kolotouros2021prohmr,sengupta2023humaniflow}.
In this work, we propose a method that learns the full posterior distribution of plausible 3D human body parameters which is thus well suited for all aforementioned applications.

\begin{figure}
\centering
\includegraphics[width=1.0\linewidth]{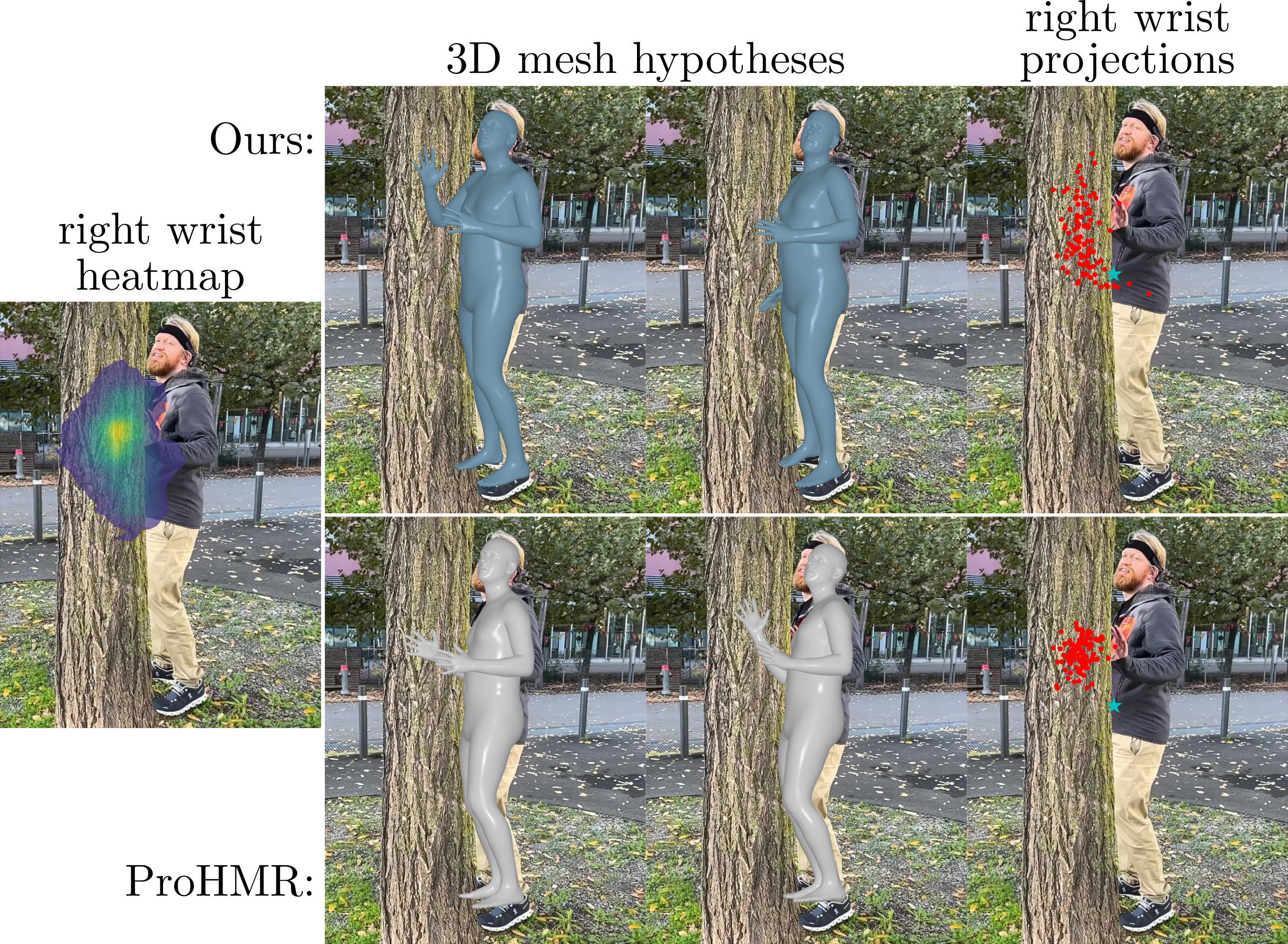}
\caption{
Our method models the full posterior distribution of plausible 3D human meshes given an RGB image.
By utilizing heatmaps of a 2D pose detector, the learned distributions have more meaningful diversity and are more accurate than distributions predicted by ProHMR~\cite{kolotouros2021prohmr}.
Two mesh hypotheses and the projection of 100 right wrist samples are shown.
$\star$ is the ground-truth 2D position of the right wrist.}
\label{fig:teaser}
\end{figure}

Recent probabilistic 3D human mesh estimation approaches~\cite{kolotouros2021prohmr,sengupta2023humaniflow,chen2023mhentropy,xu2024scorehypo,sengupta2021hierprob} employ a generative model trained by maximizing the likelihood of the ground-truth pose given an RGB image.
To further constrain the learned distribution and encourage consistency with the input, pose hypotheses are typically projected to the image and supervised with ground-truth 2D keypoints.
When only utilizing these two objective functions, learning solely depends on the availability of images with corresponding 3D human mesh annotations.
Ideally, one would have highly similar images with significantly different plausible 3D bodies, especially for examples with occlusions.
However, 3D annotations are still rather scarce and, as we argue, on its own not sufficient to best capture the conditional 3D human mesh distribution.
On the other hand, there exists an abundant amount of images with ground-truth 2D poses (\eg~\cite{lin14mscoco,wu19aichallenger,andriluka14mpii,li19crowdpose,li19ochuman}).
Wehrbein~\etal~\cite{wehrbein2021iccv} were the first to show that 2D detectors~\cite{xu22vitpose,sun19hrnet} trained on such large scale datasets encode joint occurrence probabilities in the predicted heatmaps that can be leveraged for probabilistic 3D human pose estimation.
They simplify each heatmap as a 2D Gaussian which is then used as condition and for supervision during training.
DiffPose~\cite{holmquist2023diffpose} computes a condition based on samples drawn from the heatmaps, but does not utilize them for supervising the generated distributions.
In contrast, our method directly uses the distributions encoded in heatmaps as supervision target without having to oversimplify them.
Furthermore, we go beyond modeling only the 3D positions of a small number of predefined joints.

In this work, we aim to train a probabilistic 3D human mesh estimation method that fully utilizes an off-the-shelf 2D pose detector~\cite{xu22vitpose} to best capture the full posterior distribution of plausible 3D body poses given an RGB image.
We employ normalizing flows~\cite{kingma16iaf,rezende2015nf} to model the conditional distributions.
The condition consists of an image feature vector together with the maximum-likelihood pose of a 2D detector.
We show that although the image is used as input, heatmaps provide important additional supervision signals (see Fig.~\ref{fig:teaser}).
Our key contribution is to directly supervise the learned distributions during training by minimizing the distance to the distributions encoded in the heatmaps.
This is done by generating multiple 3D hypotheses, projecting selected keypoints to the image, and comparing them joint-wise with samples drawn from the heatmaps.
As a distance measure, we use the Maximum Mean Discrepancy (MMD)~\cite{gretton12mmd}, which only requires samples from the distributions and no explicit density estimation.
Hence, arbitrary complex distributions can be reproduced, avoiding the need of oversimplifying the heatmaps as in~\cite{wehrbein2021iccv}.
By distilling the uncertainty information of a 2D detector into our 3D model, we circumvent the problem of having insufficient 3D annotations to learn the full 3D human mesh distribution.

\begin{figure}
\centering
\includegraphics[width=1.0\linewidth]{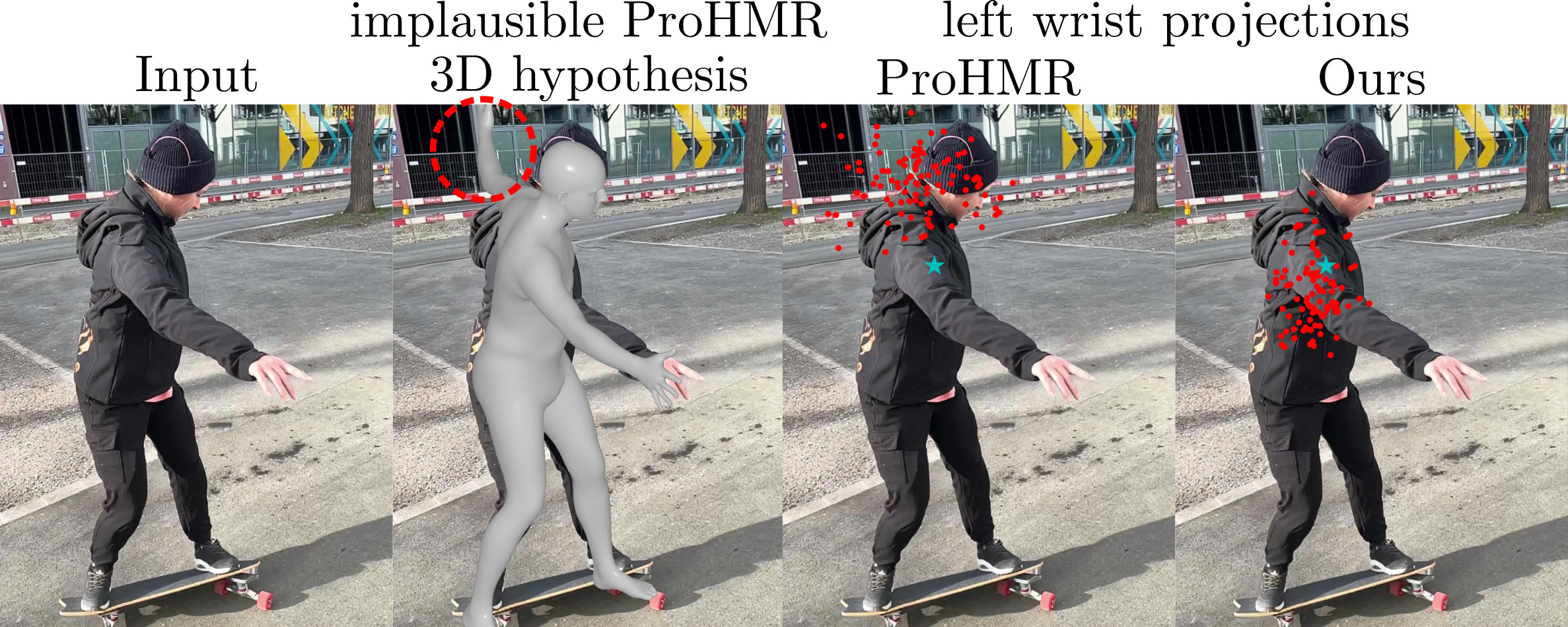}
\caption{
Multi-hypothesis 3D human pose estimation methods (\eg ProHMR~\cite{kolotouros2021prohmr}) often generate implausible hypotheses, with joints visible that should be invisible.
We significantly reduce the number of incorrect hypotheses by utilizing segmentation masks during training.
$\star$ is the ground-truth 2D position of the left wrist.}
\label{fig:mask_teaser}
\end{figure}

Our second major contribution reveals a problem of current multi-hypothesis human pose estimation approaches.
As shown in Fig.~\ref{fig:mask_teaser}, we observe that incorrect samples are often generated, where joints are visible that should be invisible.
Such failure cases are not captured in current evaluation protocols and are likely caused by the common practice of only penalizing visible joints during training.
To fill this evaluation gap, we propose two simple metrics utilizing person segmentation masks.  
The first metric measures the percentage of generated joint samples that lie inside the masks, and the second metric computes the minimum distance to the masks.
The idea is that while it is not feasible to describe all possible locations an invisible joint is allowed to be, it is clear that all joints outside the person mask are invalid.
We additionally show that by utilizing the person masks during training, we can decrease the number of invalid hypotheses generated by our model.

We evaluate our approach on 3DPW~\cite{marcard2018eccv} and EMDB~\cite{kaufmann2023emdb}, outperforming all competitors.
Our main contributions are summarized as follows:
\begin{itemize}
    \item We show that probabilistic 3D human mesh estimation benefits from predicted heatmaps of a 2D detector. 
    \item We fully utilize the heatmaps by directly using them for supervising the learned distributions of our model in a sample-wise manner.
    \item We demonstrate that person masks can be used to decrease the number of incorrect hypotheses and introduce two new metrics to evaluate it.
\end{itemize}

\section{Related Work}

This section first briefly introduces deterministic methods for monocular 3D human pose and shape estimation.
Subsequently, probabilistic approaches for reconstructing 3D human meshes and related work on probabilistic 3D human pose estimation from 2D keypoints are discussed.

\noindent\textbf{Deterministic 3D pose and shape estimation.}
Most methods output a single deterministic estimate given a monocular image.
Pioneering work~\cite{guan09iccv,hasler10cvpr,sigal07nips,bogo16eccv,lassner17up} in this area follow an optimization-based approach by fitting parameters of a body model such as SMPL~\cite{loper2015smpl,pavlakos2019smplx} to 2D image observations.
However, the optimization process tends to be slow and is sensitive to the initialization and the quality of the given image cues.
Starting with HMR~\cite{kanazawa18hmr}, learning-based approaches that directly regress the 3D body parameters from an image using a deep network became the leading paradigm~\cite{kolotouros19spin,kocabas21pare,dwivedi23poco,zhang21pymaf,wang23refit,li21hybrik,li2022cliff,dwivedi24tokenhmr,song24posturehmr,gwon24cvpr,wehrbein2023iccvw}.
They typically consist of an image backbone such as ResNet~\cite{he2016residual} or HRNet~\cite{sun19hrnet} followed by a regression head.

\noindent\textbf{Probabilistic estimation.}
The ambiguity of reconstructing 3D human pose from a monocular image was already extensively analyzed by early work~\cite{lee2004cvpr,simo2012cvpr,sminchisescu2001cvpr,sminchisescu2003cvpr,lee1985hpe,sminchisescu2002eccv,choo01iccv}.
They specify posterior distributions of 3D poses given 2D observations together with sampling strategies to generate multiple plausible solutions.
More recently, several approaches are introduced that use a deep generative model to directly learn the posterior distribution.
Li and Lee~\cite{li2019mdn} and Oikarinen~\etal~\cite{oikarinen2020graphmdn} employ a Mixture Density Network~\cite{bishop1994mdn} and generate a fixed number of 3D pose hypotheses defined by the mean of the Gaussian kernels.
Sharma~\etal~\cite{sharma2019iccv} use a variational autoencoder conditioned on 2D pose detections which is more flexible and can produce an unlimited number of hypotheses.
Since these approaches only take 2D keypoints as input, they cannot correctly model occlusions.
Addressing this problem, Wehrbein~\etal~\cite{wehrbein2021iccv} propose to utilize Gaussians fitted to heatmaps of a 2D pose detector.
They use them to condition a normalizing flow and for supervising the learned distributions during training.
Instead of oversimplifying heatmaps as Gaussians, DiffPose~\cite{holmquist2023diffpose} computes an embedding based on samples drawn from the heatmaps which is used to condition a diffusion model~\cite{ho20ddpm}.
In contrast, we do not use an embedding of heatmap samples as condition, but use the samples directly to supervise the learned distributions.

Other methods extend beyond 3D joints and aim to learn the posterior of full 3D human bodies given an image.
Biggs~\etal~\cite{biggs2020multibodies} modify HMR~\cite{kanazawa18hmr} to produce a fixed-size set of hypotheses using multiple regression heads, while Sengupta~\etal~\cite{sengupta2021prob} predict a Gaussian distribution over SMPL body pose and shape parameters.
HierProbHumans~\cite{sengupta2021prob} and ProPose~\cite{fang23propose} output a matrix Fisher distribution over body pose rotations.
To learn more expressive distributions over SMPL parameters, ProHMR~\cite{kolotouros2021prohmr} uses a normalizing flow based on the Glow architecture~\cite{kingma2018glow}.
HuManiFlow~\cite{sengupta2023humaniflow} extends ProHMR to factorize full body pose into per-body-part pose distributions in an autoregressive manner.
Other normalizing flow-based approaches are MHEntropy~\cite{chen2023mhentropy} and HuProSO3~\cite{dunkel24nfso3}.
Diffusion models~\cite{ho20ddpm} are also employed to tackle the ill-posed nature of 3D human mesh recovery from monocular images~\cite{xu2024scorehypo,foo2023hmdiff} and from egocentric view~\cite{zhang23egohmr}.
However, they cannot assign likelihoods to each sample and cannot be used as conditional prior for parametric model fitting.
None of the methods try to utilize uncertainty information encoded in the heatmaps of a 2D pose detector, nor do they address the issue of generating incorrect hypotheses for occluded body parts.

\section{Preliminaries}
\noindent\textbf{Human body representation.}
We use SMPL~\cite{loper2015smpl} to represent the 3D human body.
SMPL is a differentiable parametric model that given pose and shape parameters outputs a 3D mesh $\mathcal{M}(\boldsymbol{\theta}, \boldsymbol{\beta}) \in \mathbb{R}^{N\times3}$ with $N=6890$ vertices.
The body pose $\bm{\theta} \in \mathbb{R}^{24\times3}$ consists of 23 relative joint rotations plus a global orientation, whereas the shape parameters $\bm{\beta} \in \mathbb{R}^{10}$ are coefficients of a PCA shape space.
3D joint locations can be expressed as a linear combination of mesh vertices $\mathcal{J}_\text{3D}=W \mathcal{M} \in \mathbb{R}^{J\times3}$ using a regressor $W$.

\noindent\textbf{Normalizing flows.}
A normalizing flow (NF) is a generative model used to learn arbitrarily complex data distributions through a series of invertible mappings of a simple base distribution.
Let $Z \in \mathbb{R}^D$ be a random variable from a base distribution $p_Z(\mathbf{z})$, typically specified as $\mathcal{N}(\mathbf{0}, \mathbf{I})$, and $\mathbf{f} : \mathbb{R}^D \to \mathbb{R}^D$ an invertible and differentiable function.
The normalizing flow then transforms $Z$ into a target distribution $X = \mathbf{f}(Z)$ using the change-of-variables formula
\begin{equation}
p_X(\mathbf{x}) = p_Z(\mathbf{z})	\, \left\vert\operatorname{det}\frac{\partial \mathbf{f}(\mathbf{z})}{\partial \mathbf{z}}\right\vert^{-1}.
\end{equation}

Typically, $\mathbf{f} = \mathbf{f}_L \circ \ldots \circ \mathbf{f}_1$ is the composition of multiple functions, each implemented using a deep neural network.
To ensure that all $\mathbf{f}_l$ are invertible, specialized flow architectures have been introduced~\cite{dinh2015nice,dinh2017realnvp,kingma2018glow,germain2015made,kingma16iaf,papamakarios17flow}.
The log-probability density of the target data distribution $X$ is then defined as
\begin{equation}
\log p_X(\mathbf{x}) = \log p_Z(\mathbf{z})	-\sum_{l=1}^L \log \left\vert\operatorname{det}\frac{\partial \mathbf{f}_l(\mathbf{z}_{l-1})}{\partial {\mathbf{z}}_{l-1}}\right\vert,
\end{equation}
which can be directly optimized during training.
By introducing a condition vector $\mathbf{c} \in \mathbb{R}^{D_c}$ and using transformations $\mathbf{x}=\mathbf{f}(\mathbf{z};\mathbf{c})$, $\mathbf{f}: \mathbb{R}^D \times \mathbb{R}^{D_c} \rightarrow \mathbb{R}^D$, NFs can model conditional probability distributions $p_{X|C}(\mathbf{x} \vert \mathbf{c})$~\cite{winkler19conditionalnf}.

\begin{figure*}
\centering
\includegraphics[width=0.9\linewidth]{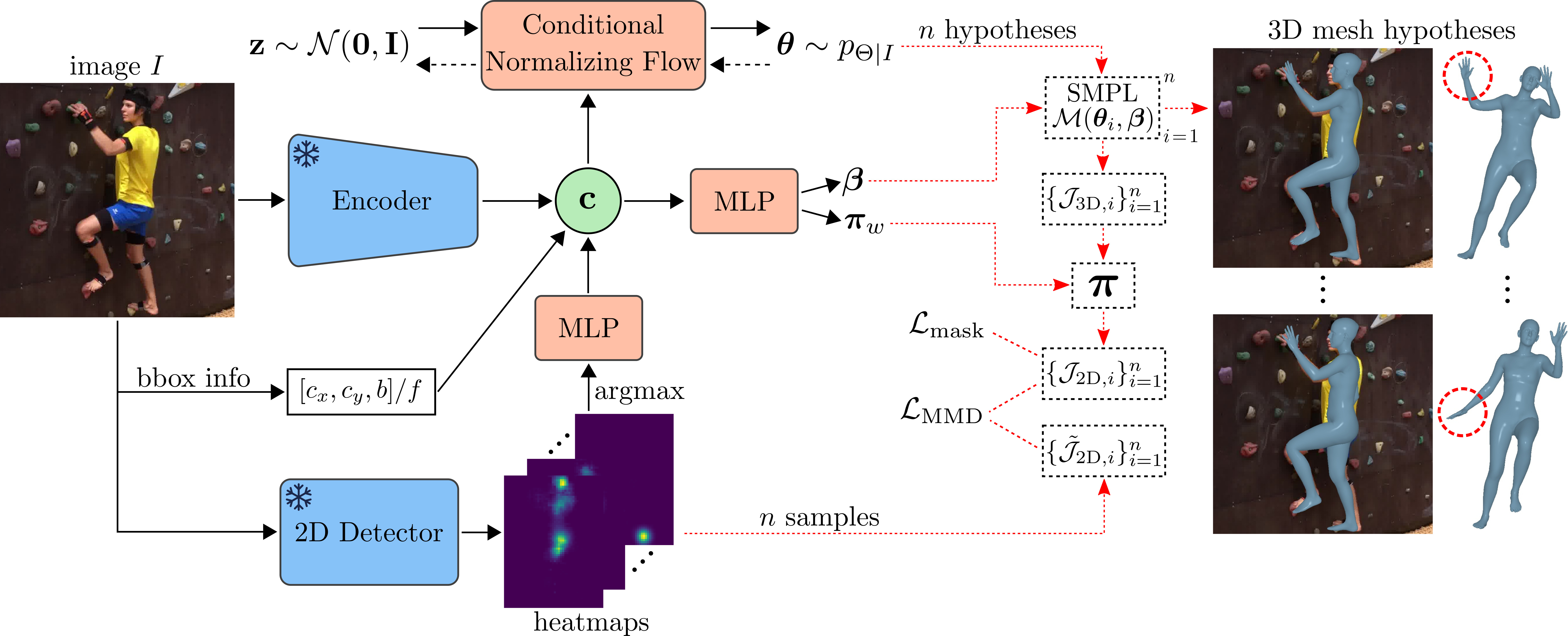}
\caption{Overview of our approach.
Given an image $I$, we model the full posterior distribution of plausible 3D human meshes using a normalizing flow. 
In addition to maximizing the likelihood of the ground-truth pose, we supervise the learned distributions by minimizing the Maximum Mean Discrepancy between heatmap samples and projections of 3D mesh hypotheses generated by our NF.
Furthermore, a segmentation mask loss is used to penalize invalid hypotheses.
Body shape $\bm{\beta}$ and camera parameters $\bm{\pi}_w$ are estimated deterministically.
}
\label{fig:method}
\vspace{-1.0em}
\end{figure*}

\section{Method}
Our goal is to learn the full posterior distribution of plausible 3D human meshes given an image.
Following ProHMR~\cite{kolotouros2021prohmr}, we model the distribution of SMPL pose parameters using a normalizing flow, while predicting deterministic estimates for shape and camera parameters (Sec.~\ref{sec:model_design}).
Unlike all previous work, we \textit{fully} utilize predicted heatmaps of a state-of-the-art 2D pose detector~\cite{xu22vitpose}, since they encode useful information about joint occurrence probabilities. 
Our model not only uses the highest-likelihood 2D pose as condition, but additionally directly supervises the learned distribution with samples from the heatmaps (Sec.~\ref{sec:heatmap_sampling}).
Furthermore, we exploit human segmentation masks to penalize incorrect hypotheses for ambiguous body parts (Sec.~\ref{sec:human_masks}).
An overview of the proposed method is shown in Fig.~\ref{fig:method}.

\subsection{Model Design}
\label{sec:model_design}
Given an image $I$ cropped around a person, we first use HRNet~\cite{sun19hrnet} as an image encoder to compute its context vector $\mathbf{c}_I \in \mathbb{R}^{720}$.
Additionally, $I$ is fed to the 2D pose detector ViTPose~\cite{xu22vitpose} which outputs one heatmap per joint, encoding the probability of a joint's presence at each location. 
The highest-likelihood 2D pose with its corresponding confidence scores is then used as input to a small MLP to produce a feature vector $\mathbf{c}_P \in \mathbb{R}^{256}$.
Since human mesh estimation methods process images cropped around a person, the location of the person in the original full-frame image is lost, introducing extra ambiguity for global rotation estimation.
To retain this information, we follow CLIFF~\cite{li2022cliff} and use the center $c_x, c_y$ and scale $b$ of the person's bounding box normalized by the focal length $f$ as feature vector:
\begin{equation}
\mathbf{c}_B = [c_x, c_y, b] / f.
\end{equation}
Finally, our condition is constructed by concatenating the three feature vectors: $\mathbf{c} = [\mathbf{c}_I, \mathbf{c}_P, \mathbf{c}_B] \in \mathbb{R}^{979}$.

We employ a conditional normalizing flow to model the posterior distribution of SMPL joint rotations $\log p_{\Theta \vert I}({\bm{\theta}} \vert \mathbf{c})$.
The architecture is based on RealNVP~\cite{dinh2017realnvp} which consists of multiple stacked affine coupling layers, each predicting element-wise scale and shift coefficients for an affine transformation.
To stabilize training, we apply a soft-clamping mechanism~\cite{ardizzone2019inn}, preventing scale components from diverging.
We found that it is important to use expressive non-volume preserving flows instead of simpler volume-preserving ones~\cite{dinh2015nice} as used in ProHMR~\cite{kolotouros2021prohmr}.

The full condition vector $\mathbf{c}$ is also utilized to regress deterministic estimates for the body shape $\bm{\beta}$ and for a weak-perspective camera model $\bm{\pi}_w = [s, t_x, t_y]$, employing the same MLP architecture as in HMR~\cite{kanazawa18hmr,kolotouros2021prohmr}.
Using either the ground-truth or an approximated focal length~\cite{kissos2020weakpersp}, we transform the weak-perspective projection to perspective projection parameters $\bm{\pi}_p$ of the original image~\cite{li2022cliff}.

\subsection{Sampling from Heatmaps}
\label{sec:heatmap_sampling}
Most 2D pose estimators are optimized to regress one heatmap per joint, with ground-truth heatmaps consisting of a 2D Gaussian centered at the joint location. 
As a by-product of training, heatmaps of a strong detector represent the confidence (\ie a value between $0$ and $1$) of the corresponding keypoint being at any location.
A heatmap is therefore a finite set of non-negative values, and can hence be interpreted as a multinomial distribution with $48 \times 64$ possible outcomes (the output dimensions of ViTPose~\cite{xu22vitpose}).
We independently draw $n$ samples per joint from the constructed distributions.
Instead of using an embedding of the samples as condition~\cite{holmquist2023diffpose}, we directly use them for supervising the learned distributions of our normalizing flow.
To this end, we draw $n$ SMPL pose hypotheses from $p_{\Theta \vert I}({\bm{\theta}} \vert \mathbf{c})$ and project the target keypoints to the image using the regressed camera $\bm{\pi}_p$.
For joint $k$, given heatmap samples $\hat{\mathcal{S}}_k = {\{\hat{\mathbf{s}}_{k,i}}\}^n_{i=1}$ and projected NF samples $\mathcal{S}_k = {\{\mathbf{s}_{k,i}}\}^n_{i=1}$ both normalized to the range \mbox{[-1, 1]}, we compute the Maximum Mean Discrepancy with kernel $\varphi$ as
\begin{equation}
\label{eq:mmd}
 \begin{aligned}
\mathcal{L}&_{\text{MMD}}(\mathcal{S}_k, \hat{\mathcal{S}}_k)=\frac{1}{n(n-1)} \sum_{i \neq j}^{n} \varphi\left(\mathbf{s}_{k,i}, \mathbf{s}_{k,j}\right) \\ &+ \frac{1}{n(n-1)} \sum_{i \neq j}^{n} \varphi\left(\hat{\mathbf{s}}_{k,i}, \hat{\mathbf{s}}_{k,j}\right) - \frac{2}{n^2} \sum_{i,j=1}^{n} \varphi\left(\mathbf{s}_{k,i}, \hat{\mathbf{s}}_{k,j}\right).
\end{aligned}
\end{equation}
Note that we calculate the MMD joint-wise and not pose-wise, because the heatmaps encode the occurrence probabilities of joints independently.
Since the detector tends to overestimate the diversity of clearly visible keypoints, we only use heatmap samples for uncertain joints, and $n$ times duplicated ground-truth positions for the certain ones.

\subsection{Penalizing Incorrect Hypotheses}
\label{sec:human_masks}
\begin{figure}
\centering
\includegraphics[width=0.85\linewidth]{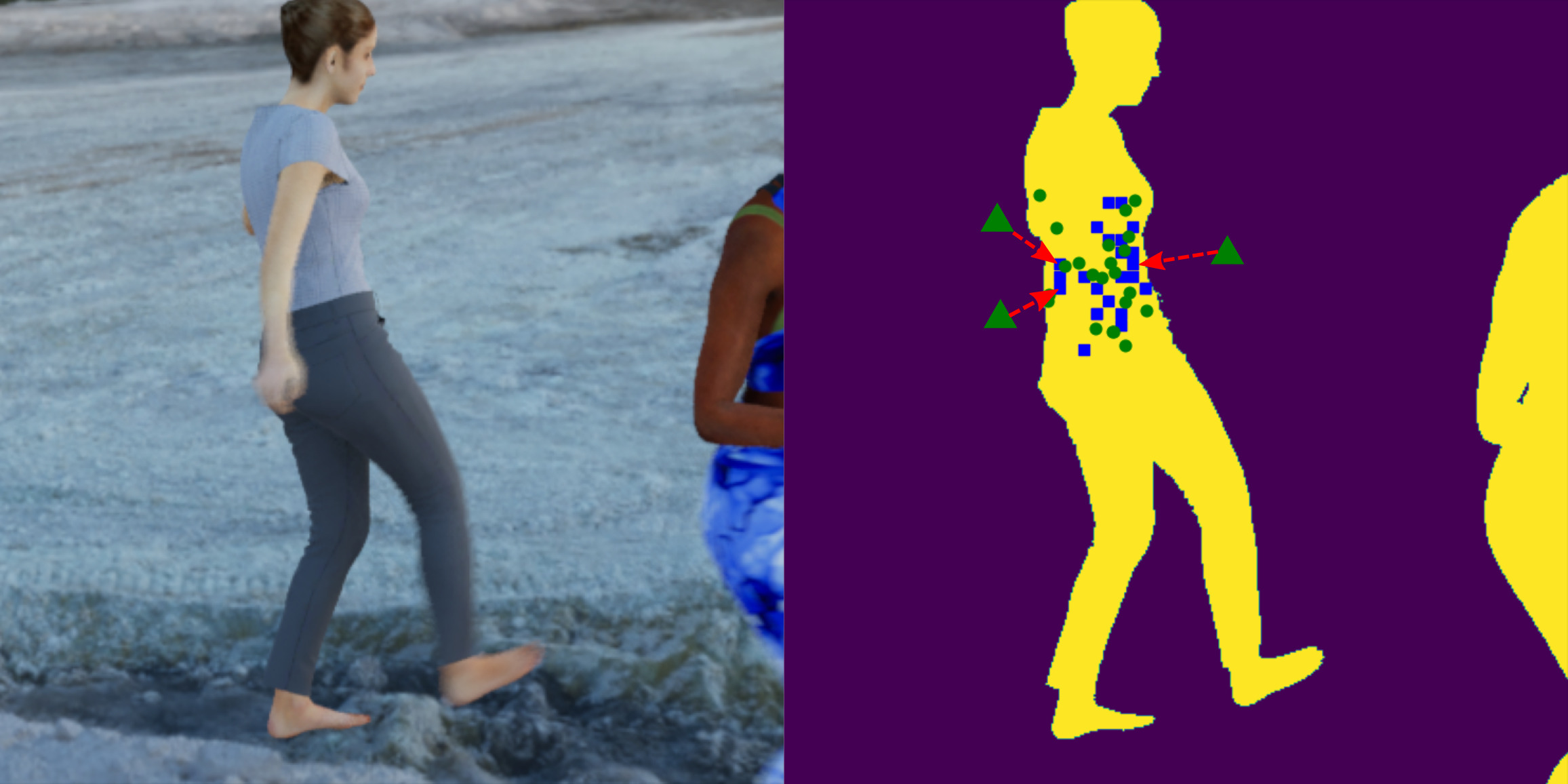}
\caption{Visualization of the person mask loss $\mathcal{L}_{\text{mask}}$ for the left wrist of the person in the center.
We explicitly penalize hypotheses for invisible joints that lie outside the person masks by minimizing the $l_1$ distance to the closest corresponding heatmap samples.
Plausible hypotheses not penalized by $\mathcal{L}_{\text{mask}}$ are shown as green dots, implausible ones as green triangles and heatmap samples as blue squares.
Best viewed with zoom and in color.}
\label{fig:mask_loss}
\end{figure}

We aim to produce diverse 3D hypotheses, especially for occluded body parts.
However, we notice that the diversity produced for invisible joints is sometimes too large, as evidenced by samples generated that are visible (see Fig.~\ref{fig:mask_teaser}).
To penalize these incorrect hypotheses, we propose to utilize person segmentation masks.
Intuitively, 3D hypotheses that project outside the person mask are misaligned with the image evidence and thus incorrect.
As visualized in Fig.~\ref{fig:mask_loss}, we compute an $l_1$ loss, $\mathcal{L}_{\text{mask}}$, between projected 2D joints of NF samples that lie outside the person mask and the closest corresponding heatmap samples.
Note that this is more efficient than optimizing the distance to the closest mask pixel and, as we observe, leads to more robust training.
We remove outside of mask heatmap samples before computing the loss and find this more effective than removing these samples for MMD computation (Eq.~\ref{eq:mmd}).
Unlike a typical 2D reprojection loss which minimizes the distance to the corresponding ground-truth keypoint, our mask loss $\mathcal{L}_{\text{mask}}$ does not falsely restrict the learned distributions and is hence suitable for occluded joints.

\subsection{Optimization}
Given an image with corresponding ground-truth SMPL parameters, we minimize the negative log-likelihood of the pose $\hat{\bm{\theta}}$ and apply an MSE loss for the shape $\hat{\bm{\beta}}$:
\begin{equation}
\begin{aligned}
\mathcal{L}_{\text{NLL}} &= - \log p_{\Theta \vert I}(\hat{\bm{\theta}} \vert \mathbf{c}), \\
\mathcal{L}_{\beta} &= \|\bm{\beta} - \hat{\bm{\beta}}\|_2^2.
\end{aligned}
\end{equation}
Additionally, we compute an $l_1$ loss between 2D ground-truth keypoints, $\hat{\mathcal{J}}_\text{2D}$, and projected keypoints of the approximated mode of the output distribution $\bm{\theta}^*$, corresponding to the NF sample with all-zeros latent vector $\mathbf{z} = \mathbf{0}$:
\begin{equation}
\mathcal{L}_\text{2D} = \|\bm{\pi}_p(W\mathcal{M}(\bm{\theta}^*,\bm{\beta})) - \hat{\mathcal{J}}_\text{2D}\|_1.
\end{equation}
Note that we apply this loss only to the approximated mode prediction and not to random samples, as this helps producing better single-hypothesis estimates while not unnecessarily constraining the modeled distribution.
Finally, since we use the 6D rotation representation of~\cite{zhou19rot6d} in our NF, we regularize generated samples to be orthonormal with $\mathcal{L}_{\text{orth}}$ following ~\cite{kolotouros2021prohmr,zhang23egohmr}. 
With the proposed MMD (Sec.~\ref{sec:heatmap_sampling}) and mask loss (Sec.~\ref{sec:human_masks}), the overall loss is then defined as:
\begin{equation}
\begin{aligned}
    \mathcal{L} &= \lambda_{\beta}\mathcal{L}_{\beta} + \lambda_{\text{2D}}\mathcal{L}_{\text{2D}} + \lambda_{\text{NLL}}\mathcal{L}_{\text{NLL}} \\
    &+ \lambda_{\text{orth}}\mathcal{L}_{\text{orth}} + \lambda_{\text{MMD}}\mathcal{L}_{\text{MMD}} + \lambda_{\text{mask}}\mathcal{L}_{\text{mask}},
\end{aligned}
\end{equation}
where $\lambda$ are the corresponding weight coefficients.

\subsection{Implementation Details}
\noindent\textbf{2D detector and heatmap sampling.} We employ the state-of-the-art model ViTPose-H~\cite{xu22vitpose} with the publicly available checkpoint trained on multiple human datasets.
To reduce training time, we precompute ViTPose predictions with the ground-truth image crop and save 150 heatmap samples per joint using the 17 keypoints COCO skeleton~\cite{lin14mscoco} together with their minimum distance to the person mask.
Samples with confidence smaller than $0.05$ are discarded to suppress outliers.
Since in 3D human mesh estimation, the root joint (pelvis) is typically centered in the crop and relative rotations are estimated, the possible variance of joints close to the root of the kinematic tree is limited.
We found that \mbox{ViTPose} often generates distributions with too much diversity for these joints, as well as for the facial landmarks.
Therefore, we only use heatmap samples for joints that are highly articulated (\ie elbows, wrists, knees and ankles) and uncertain.
We define a keypoint as being uncertain if its corresponding ViTPose heatmap has a maximum confidence value below $0.7$, which typically happens due to \eg occlusion, motion blur or unusual clothing.
Furthermore, we use the maximum confidence values as proxy for the visibility of joints, and set the threshold to $0.5$.
For non-highly articulated joints, ground-truth positions are used and a prerequisite for applying $\mathcal{L}_{\text{MMD}}$ is that the joint is visible.
Finally, in order to not falsely restrict the learned distributions, $\mathcal{L}_{\text{MMD}}$ is only applied if the ground-truth joint is inside the image crop.
We randomly draw $n = 25$ samples and use a mixture of inverse multiquadratics kernels~\cite{tolstikhin18wasserstein,ardizzone2019inn}
\begin{equation}
\varphi^{im}_{\mathcal{B}}(\mathbf{s}, \hat{\mathbf{s}})=\sum_{a \in \mathcal{B}} \frac{a^2}{a^2+\left\|\mathbf{s}-\hat{\mathbf{s}}\right\|^2}
\end{equation}
with bandwidth parameters $\mathcal{B} = \{0.05, 0.20, 0.90\}$~\cite{ardizzone2019inn}.

\noindent\textbf{Person segmentation masks} can be either obtained from strong segmentation models (\eg~\cite{ravi2024sam2,ghiasi21seg,he2017maskrcnn,khirodkar2024sapiens}) or are already provided in synthetic human pose datasets~\cite{black2023bedlam,patel2021agora}.
To penalize and later on evaluate incorrect hypotheses, it is crucial that the person mask contains at least every pixel of the image where the target person could be.
However, this is not the case if the person is occluded by another person or by an object, because the mask only contains visible person pixels.
Thus, as shown in Fig.~\ref{fig:mask_loss}, we modify the segmentations by taking the union of all person masks in an image.
We detect object occlusions by projecting the ground-truth body to the image and measuring the overlap with the corresponding mask.
The mask loss $\mathcal{L}_{\text{mask}}$ is then not applied to these examples, since it would be impractical to generate segmentations for every possible objects occluding a person.
We define a ground-truth joint to be invisible if its location is outside the image crop or its corresponding maximum confidence score is below $0.5$.

\noindent\textbf{Network and training details.} 
As image backbone, we use an HRNet-W48~\cite{sun19hrnet} with pretrained checkpoint from BEDLAM-CLIFF~\cite{black2023bedlam}.
We freeze the weights as this significantly speeds up training while resulting in similar performance.
Since the training data contains images with heavily overlapping people, ViTPose sometimes computes estimates for a person other than the target.
We filter out these examples by thresholding the Euclidean distance between ground-truth and ViTPose 2D poses.
Further implementation and training details are provided in the supplementary.

\section{Experiments}

\begin{table*}
	\centering
 \resizebox{0.8\textwidth}{!}{
	\begin{tabular}{clccc|ccc}
		\toprule
        & & & 3DPW (14) & & & EMDB (24) \\
		\cmidrule(lr){3-5}\cmidrule(lr){6-8}
		  & Models &
		\multicolumn{1}{c}{MPJPE $\downarrow$} & \multicolumn{1}{c}{PA-MPJPE $\downarrow$} & \multicolumn{1}{c}{PVE $\downarrow$} & \multicolumn{1}{c}{MPJPE $\downarrow$} & \multicolumn{1}{c}{PA-MPJPE $\downarrow$} & \multicolumn{1}{c}{PVE $\downarrow$}  \\
		\midrule
        \parbox[t]{4mm}{\multirow{8}{*}{\rotatebox[origin=c]{90}{ Best of 1}}} & Biggs \etal \cite{biggs2020multibodies} NeurIPS'20  & 93.8 & 59.9 & - & - & - & - \\
        & Sengupta \etal \cite{sengupta2021prob} CVPR'21  & 97.1 & 61.1 & - & - & - & - \\
		  & ProHMR~\cite{kolotouros2021prohmr} ICCV'21 & 97.0 & 59.8 & - & - & - & - \\
        & HierProbHuman~\cite{sengupta2021hierprob} ICCV'21 & 84.9 & 53.6 & - & - & - & -\\
        & HuManiFlow*~\cite{sengupta2023humaniflow} CVPR'23  & 83.1 & 53.9 & 98.6 & 113.9 & 76.4 & 133.0   \\
        & BEDLAM-CLIFF~\cite{black2023bedlam} CVPR'23 & 66.9 & 43.0 & 78.5 & 98.0 & 60.6 & 111.6 \\
        & ScoreHypo~\cite{xu2024scorehypo} CVPR'24  & 72.4 & 44.5 & 84.6 & 112.4* & 77.9* & 131.5* \\
        & ProHMR\textsuperscript{\textdagger}~\cite{kolotouros2021prohmr}  & 71.7 & 43.5 & 84.7 & 98.2 & 60.4 & 114.5\\
        &\cellcolor{Gray} Ours &\cellcolor{Gray} $\bm{62.2}$ &\cellcolor{Gray} $\bm{40.9}$ &\cellcolor{Gray} $\bm{73.9}$ &\cellcolor{Gray} $\bm{84.3}$ &\cellcolor{Gray} $\bm{56.1}$ &\cellcolor{Gray} $\bm{97.7}$ \\
        \hline
        \parbox[t]{4mm}{\multirow{8}{*}{\rotatebox[origin=c]{90}{Best of 100}}} & Biggs \etal \cite{biggs2020multibodies} NeurIPS'20  & 74.6 & 48.3 & - & - & - & - \\
        & Sengupta \etal \cite{sengupta2021prob} CVPR'21  & 84.4 & 52.1 & - & - & - & - \\
        & ProHMR~\cite{kolotouros2021prohmr} ICCV'21  & 81.5 & 48.2 & -  & - & - & - \\
        & HierProbHuman~\cite{sengupta2021hierprob} ICCV'21  & 70.9 & 43.8 & - & - &- & -\\
        & HuManiFlow*~\cite{sengupta2023humaniflow} CVPR'23  & 64.5 & 40.0 & 75.5 & 88.7 & 56.5 & 100.6   \\
        & ScoreHypo~\cite{xu2024scorehypo} CVPR'24  & 63.0 & 37.6 & 73.4 & 87.4* & 58.5* & 99.6*\\
        & ProHMR\textsuperscript{\textdagger}~\cite{kolotouros2021prohmr}  & 55.0 & 34.7 & 65.7 & 76.7 & 47.1 & 87.3 \\
        & \cellcolor{Gray} Ours  & \cellcolor{Gray}$\bm{46.2}$ &\cellcolor{Gray} $\bm{29.8}$ &\cellcolor{Gray} $\bm{54.4}$ &\cellcolor{Gray} $\bm{63.6}$ &\cellcolor{Gray} $\bm{40.9}$ &\cellcolor{Gray} $\bm{72.0}$ \\
		\bottomrule
	\end{tabular}
    }
 	\caption{Comparison of the distribution accuracy of probabilistic 3D human mesh estimation methods on 3DPW~\cite{marcard2018eccv} and EMDB~\cite{kaufmann2023emdb}. 
    Our model achieves state-of-the-art performance for both single hypothesis and best of 100 hypotheses 3D pose and mesh metrics.
    Rows or numbers marked with * are computed using publicly available checkpoints, and \textdagger\>denotes a retrained baseline.
    Parenthesis denotes the number of body joints used to calculate MPJPE and PA-MPJPE.
  }
  	\label{table:main_results}
\end{table*}

\subsection{Datasets and Evaluation Metrics}
Following BEDLAM-CLIFF~\cite{black2023bedlam}, we train our model on BEDLAM~\cite{black2023bedlam}, AGORA~\cite{patel2021agora} and 3DPW~\cite{marcard2018eccv}.
Since BEDLAM and AGORA are synthetic datasets, they provide high quality person segmentation masks which we utilize during training.
We do not apply our mask loss to examples from 3DPW, as computing accurate masks for the crowded scenes would incur a large overhead, while 3DPW only accounts for a small fraction of the total training data.

We evaluate on the challenging in-the-wild datasets 3DPW and EMDB (subset 1)~\cite{kaufmann2023emdb}, and follow HuManiFlow~\cite{sengupta2023humaniflow} to assess the \textit{accuracy}, input \textit{consistency} and \textit{diversity} of the estimated 3D human mesh distributions.
The distribution accuracy is measured by calculating the Mean Per Joint Position Error (MPJPE), the MPJPE after Procrustes Alignment (PA-MPJPE) and the Per Vertex Error (PVE), all in $mm$.
We report these metrics for the best 3D hypothesis generated by the predicted distribution, evaluating whether the ground-truth is contained in the distribution.
To validate that all hypotheses are consistent with the input image, we compute the Euclidean distance between \textit{visible} ground-truth 2D keypoints (2DKP) and the corresponding reprojected predicted samples, averaged over 100 hypotheses per image.
Following \cite{sengupta2023humaniflow}, the 17 keypoints COCO skeleton~\cite{lin14mscoco} is used and the 2DKP error is calculated in $256 \times 256$ pixel space.
ViTPose confidence values with a threshold of $0.5$ are used as proxy for the visibility of keypoints.
The distribution diversity is measured by generating 100 3D keypoint (3DKP) hypotheses, and computing the average Euclidean distance to the mean for each keypoint in $mm$.
This is split into visible and invisible keypoints, expecting higher diversity for invisible ones.
Since for invisible joints, it is not meaningful to compute the 2DKP error and a high diversity is even desired, hypotheses misaligned with the image evidence are currently not captured by the evaluation protocols.
Thus, we introduce two metrics utilizing person segmentation masks which measure the \textit{plausibility} of hypotheses generated for invisible joints.
We define an invisible joint as plausible if it is inside the person mask, and compute the percentage of hypotheses inside the mask (PercIn) and the minimum Euclidean distance to the mask (MinDist). 
Mask-RCNN~\cite{ghiasi21seg} is employed to generate the segmentations.
To ensure that all masks are of high quality, we evaluate the mask metrics only on a subset of EMDB.
This subset is constructed by first selecting all images with at least two invisible keypoints and then manually filtering for quality, resulting in 1760 person masks.
EMDB is used for evaluation instead of 3DPW since it contains less cluttered scenes leading to more accurate mask estimates.

\subsection{Quantitative Evaluation}
Since models are often trained on different data and with different backbones, we establish a baseline by training ProHMR~\cite{kolotouros2021prohmr} on the same three datasets using the same image backbone and denote this competitor as ProHMR\textsuperscript{\textdagger}.
Unfortunately, we could not successfully reproduce the training of HuManiFlow~\cite{sengupta2023humaniflow} and thus use the provided checkpoint for evaluation (marked by *).
All other numbers are taken from the respective papers.
To evaluate the distribution accuracy of the probabilistic 3D human mesh estimation methods, we calculate the minimum MPJPE, PA-MPJPE and PVE of 100 hypotheses.
Additionally, we report results for a single deterministic estimate which we generate by using an all-zeros latent vector for the normalizing flow, corresponding to the approximated mode of the distribution.
The results of our method and the state-of-the-art competitors are shown in Table~\ref{table:main_results}.
We significantly outperform all competitors in all metrics on both 3DPW and EMDB.
Our method additionally yields better performance for the conventional single prediction regression task than BEDLAM-CLIFF~\cite{black2023bedlam} which is also trained on the same data with the same backbone.

\begin{table}
	\centering
 \resizebox{\linewidth}{!}{
	\begin{tabular}{clccc}
		\toprule
        & & \textit{Consistency} & \textit{Diversity} & \textit{Plausibility}\\
        & & 2DKP Error $\downarrow$ & 3DKP Spread &  Mask Metrics\\
		  &  Models &
		\multicolumn{1}{c}{Mode / Samples} & \multicolumn{1}{c}{Vis. / Invis.} &  \multicolumn{1}{c}{PercIn$\uparrow$ / MinDist$\downarrow$}   \\
		\midrule
		\parbox[t]{1mm}{\multirow{4}{*}{\rotatebox[origin=c]{90}{3DPW}}} & HF*~\cite{sengupta2023humaniflow} & 5.73 / 7.26 & 42.0 / 91.7 & -\\
        & PHMR\textsuperscript{\textdagger}~\cite{kolotouros2021prohmr} & 6.91 / 9.22 & 57.0 / 74.0 &- \\
        & Ours w/o $\mathcal{L}_{\text{mask}}$ & 4.79 / 5.96 & 35.3 / 80.0 & -\\
        & Ours w/ $\mathcal{L}_{\text{mask}}$ & $\bm{4.77}$ / $\bm{5.94}$ & 34.8 / 75.1 & -  \\
        \hline
		\parbox[t]{1mm}{\multirow{4}{*}{\rotatebox[origin=c]{90}{EMDB}}} & HF*~\cite{sengupta2023humaniflow} & 7.54 / 8.96 & 47.6 / 88.3 & 85.1 / 1.576\\
        & PHMR\textsuperscript{\textdagger}~\cite{kolotouros2021prohmr} & 8.59 / 10.94 & 62.5 / 85.9 & 86.0 / 1.334\\
        & Ours w/o $\mathcal{L}_{\text{mask}}$ & 5.03 / 6.37 & 41.0 / 94.9 & 88.8 / 1.049 \\
        & Ours w/ $\mathcal{L}_{\text{mask}}$ & $\bm{5.02}$ / $\bm{6.34}$ & 40.8 / 87.8 & $\bm{91.4}$ / $\bm{0.616}$ \\
        
		\bottomrule
	\end{tabular}
    }
 	\caption{Comparison of the input consistency, sample diversity and sample plausibility of recent probabilistic 3D human mesh estimation methods.
  By employing our mask loss, we significantly improve the sample plausibility.}
  	\label{table:distribution_eval}
\end{table}

\begin{figure*}
\centering
\includegraphics[width=0.95\linewidth]{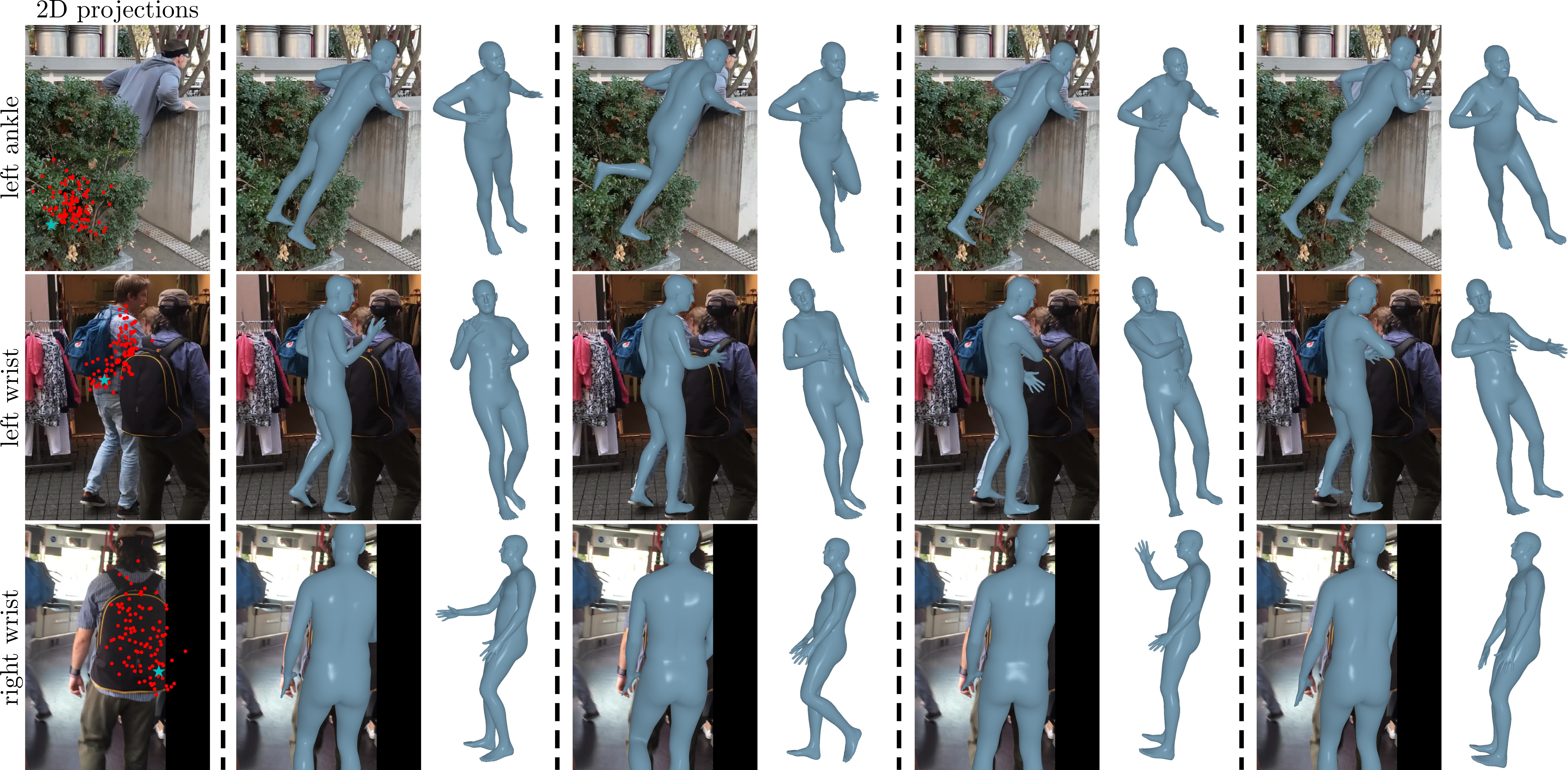}
\caption{Qualitative results for challenging in-the-wild images with significant occlusions or truncations of body parts.
Four samples from the learned 3D human mesh distribution are shown together with the 2D projections of 100 hypotheses for a highly ambiguous joint. 
}
\label{fig:results}
\end{figure*}

Table~\ref{table:distribution_eval} compares the input consistency, sample diversity and sample plausibility of recent probabilistic methods.
We achieve significantly better input consistency for the mode and for random hypotheses than the competitors. 
Furthermore, our generated 3D diversity is lowest for visible joints while being reasonable high for invisible ones.
The ProHMR baseline fails to produce meaningful diversity, evidenced by a relatively small difference in 3D spread for visible and invisible keypoints.
While HuManiFlow achieves reasonable diversity, its generated hypotheses are often incorrect as evaluated by the introduced mask metrics.
By employing our proposed mask loss $\mathcal{L}_{\text{mask}}$, the percentage of plausible hypotheses increases, and the minimum mask distance decreases significantly.
Additionally, $\mathcal{L}_{\text{mask}}$ leads to a lower 3D spread for invisible joints, while the distribution accuracy does not deteriorate (see Table~\ref{table:ablations}), showing that only unnecessary diversity is removed.
Overall, our learned 3D mesh distributions have by far the best characteristics \textit{w.r.t.}\@ accuracy, consistency, diversity and plausibility.
Note that the HuManiFlow numbers for 3DPW in Table~\ref{table:distribution_eval} slightly differ from their reported numbers since they use wider bounding boxes for 2DKP error computation and a different 2D pose detector to determine if a joint is visible.

Qualitative results are shown in Fig.~\ref{fig:results}.
Our method generates diverse and plausible configurations for ambiguous body parts, with all hypotheses consistent with the image evidence.
A wider variety of results and comparisons with competitors can be found in the supplementary material.

\subsection{Ablation Study}
\begin{figure}
\centering
\includegraphics[width=1.0\linewidth]{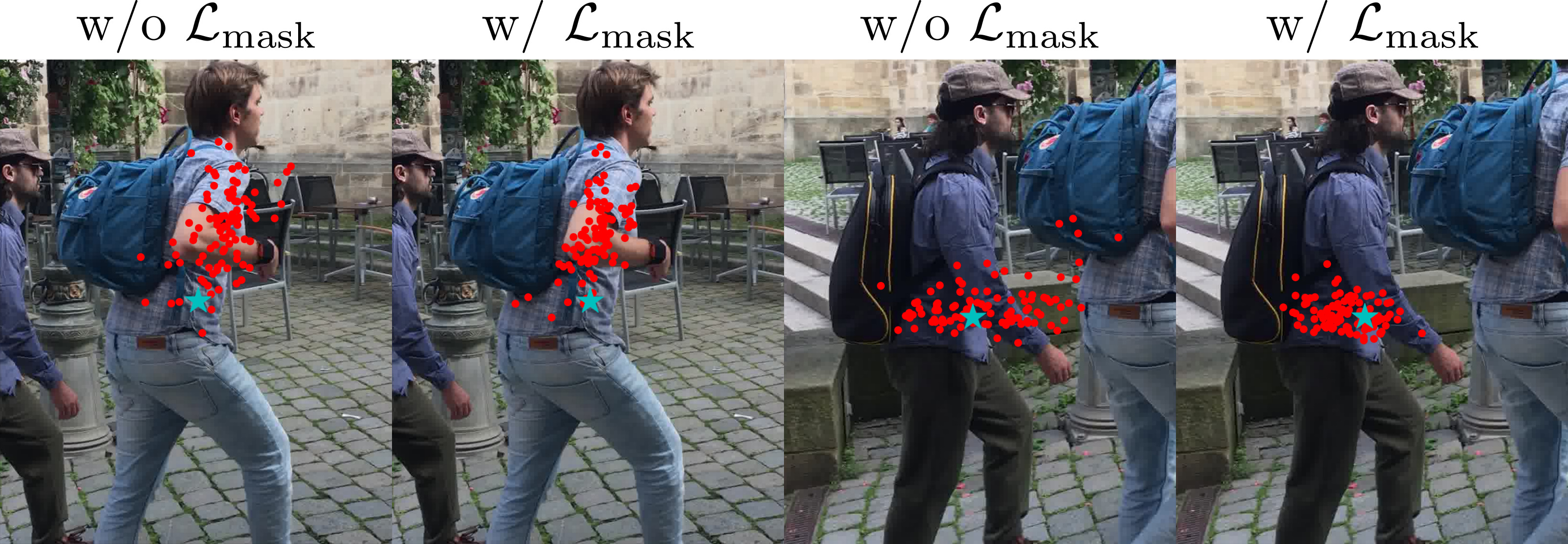}
\caption{
By using our mask loss $\mathcal{L}_{\text{mask}}$, significantly fewer incorrect hypotheses are generated while meaningful diversity is maintained.
The 2D projections of 100 hypotheses for the left wrist are shown, together with the ground-truth position ($\star$).}
\label{fig:mask_result}
\vspace{-0.5em}
\end{figure}

To investigate the influence of our proposed design choices and loss functions, we start with the baseline ProHMR\textsuperscript{\textdagger} and successively add all components.
The distribution accuracy metrics for 100 hypotheses per image evaluated on 3DPW are presented in Table~\ref{table:ablations}.
We extend the findings of CLIFF~\cite{li2022cliff} to the probabilistic domain, showing that bounding box information of the person crop is not only beneficial for deterministic regression, but also when used as condition for probabilistic human mesh modeling.
Leveraging the most likely 2D pose from ViTPose as additional condition for our NF leads to further improvements after filtering out incorrect predictions.
The non-volume preserving flow RealNVP is able to better model the distributions than the simpler volume-preserving flow NICE~\cite{dinh2015nice} used by ProHMR.
We find that NICE especially struggles with predicting sharp distributions for unambiguous joints, while estimating diverse hypotheses for invisible body parts.
By directly supervising ($\mathcal{L}_{\text{MMD}}$) the learned distributions with distributions encoded in heatmaps of \mbox{ViTPose}, the model learns to generate more meaningful diversity.
Finally, while the proposed mask loss $\mathcal{L}_{\text{mask}}$ has little influence on the 3D pose and mesh metrics, it leads to a significant decrease in the number of incorrect hypotheses and thus to higher plausibility of the estimated distributions.
This is qualitatively shown in Fig.~\ref{fig:mask_result} and quantitatively evaluated in Table~\ref{table:distribution_eval}.

\begin{table}
	\centering
 \resizebox{0.9\linewidth}{!}{
	\begin{tabular}{lccccc}
		\toprule
       & & 3DPW (14) \\
		\cmidrule(lr){2-6}
		  Models &
		\multicolumn{1}{c}{MPJPE $\downarrow$} & \multicolumn{1}{c}{PA-MPJPE $\downarrow$} & \multicolumn{1}{c}{PVE $\downarrow$} & \multicolumn{1}{c}{} & \multicolumn{1}{c}{}  \\
		\midrule
        ProHMR\textsuperscript{\textdagger}~\cite{kolotouros2021prohmr} & 55.0 & 34.7 & 65.7\\
        + bbox info~\cite{li2022cliff} & 52.3 & 34.5 & 62.6 \\
        + 2D pose condition & 49.5 & 32.7 & 58.6\\
        + RealNVP & 48.1 & 31.8 &56.9 \\
        + $\mathcal{L}_{\text{MMD}}$ & 46.5 & 29.7 & 54.8\\
        + $\mathcal{L}_{\text{mask}}$ (Ours full) & 46.2 & 29.8 & 54.4\\
		\bottomrule
	\end{tabular}
    }
 	\caption{Ablation study analyzing our proposed design choices and loss functions.
  Components are added successively, and the minimum errors out of 100 hypotheses are reported.
  }
  	\label{table:ablations}
\end{table}

\section{Conclusion}
We present a probabilistic approach to the highly ill-posed problem of monocular 3D human mesh reconstruction.
Our normalizing flow-based method utilizes joint occurrence probabilities encoded in heatmaps of a 2D pose detector as supervision target for the learned distributions.
This is accomplished by minimizing the Maximum Mean Discrepancy between samples drawn from the heatmaps and hypotheses generated by our model, and leads to more meaningfully diverse distributions.
Additionally, we reveal that current methods suffer from generating incorrect solutions for invisible joints and propose a simple loss based on segmentation masks that improves plausibility of the hypotheses. 
Given a monocular image, our method produces plausible 3D human mesh hypotheses that are consistent with the image evidence while maintaining high diversity for ambiguous body parts.
Future work could extend our approach to the temporal domain to reconstruct temporally consistent motions for sequences with strong occlusions. 

\vspace{-1.0em}
\small{\paragraph{Acknowledgements.}
This work was supported by the Federal Ministry of Education and Research (BMBF), Germany, under the AI service center KISSKI (grant no.\ 01IS22093C), the Deutsche Forschungsgemeinschaft (DFG) under Germany’s Excellence Strategy within the Cluster of Excellence PhoenixD (EXC 2122), the Wallenberg Artificial Intelligence, Autonomous Systems and Software Program (WASP), funded by Knut and Alice Wallenberg Foundation, and the European Union within the Horizon Europe research and innovation programme under grant agreement no.\ 101136006 – XTREME.
}

{\small
\bibliographystyle{ieee_fullname}
\bibliography{egbib}
}
\clearpage 

\renewcommand\thesection{\Alph{section}}
\renewcommand\thesubsection{\thesection.\arabic{subsection}}
\renewcommand{\thefigure}{S\arabic{figure}}
\renewcommand{\thetable}{S\arabic{table}}
\setcounter{figure}{0}
\setcounter{table}{0}
\setcounter{section}{0}

\section{Implementation Details}
\paragraph{Network and training.}
The normalizing flow (NF) consists of eight RealNVP~\cite{dinh2017realnvp} coupling layers, each parameterized by an MLP with three linear layers of 1024 hidden dimensions and ReLU activations in between.
The NF implementation is based on the \texttt{FrEIA} package~\cite{freia} and the soft-clamping parameter is set to $\alpha = 2.0$.
Our model is trained for 400K iterations using Adam~\cite{kingma15adam} with weight decay and learning rate set to $1e^{-4}$, and a batch size of 64.
Training takes around two days on a single A100 GPU.
We use an input image size of $224 \times 224$ and apply data augmentation following~\cite{black2023bedlam} which includes random crops, scale and different kinds of image blur, compression, and brightness modifications.
The loss weights are set to \mbox{$\lambda_{\beta} = 5e^{-4}$}, $\lambda_{\text{2D}} = 1e^{-2}$, $\lambda_{\text{NLL}} = 1e^{-1}$, $\lambda_{\text{orth}} = 1e^{-1}$, $\lambda_{\text{MMD}} = 5e^{-2}$, $\lambda_{\text{mask}} = 1e^{-1}$.

When using crop or scale data augmentation during training, it would be intuitive to apply it to the 2D pose condition as well by masking (\ie setting to zero) the corresponding keypoints.
However, we found it is beneficial to always use the highest-likelihood 2D pose of the original crop as condition.
This leads to better generalization, since the model learns to focus more on the 2D pose instead of solely on the image feature.

Since annotations for BEDLAM~\cite{black2023bedlam} were initially only released in SMPL-X~\cite{pavlakos2019smplx} format, we follow BEDLAM-CLIFF~\cite{black2023bedlam} and predict the first 22 body pose parameters of SMPL-X.
Hence, our normalizing flow models a distribution of 132 dimensions.
We use 11 shape components in the gender-neutral shape space.
The SMPL-X labels for the training set of 3DPW~\cite{marcard2018eccv} are provided by \cite{black2023bedlam}.
All evaluation is performed using the SMPL~\cite{loper2015smpl} body, by converting predicted SMPL-X meshes to SMPL using a vertex mapping $V \in \mathbb{R}^{10475\times6890}$~\cite{pavlakos2019smplx}.

\paragraph{Competitors.}
Since ScoreHypo~\cite{xu2024scorehypo} does not evaluate on EMDB~\cite{kaufmann2023emdb}, we use their released inference code to calculate the distribution accuracy metrics on EMDB in Table~1 of the main paper.
They employ VirtualPose~\cite{su2022virtualpose} to estimate the root joint depth which is required to transform their predicted 2.5D pose representations to metric 3D space.
However, we find that in rare cases VirtualPose fails to predict reasonable depth for the target person or even fails to detect the person at all, resulting in degenerated ScoreHypo outputs.
We use the predictions of neighboring frames to fill in missing estimates.
Due to the failure cases of VirtualPose, other methods to recover metric scale such as the bone-length optimization method from Pavlakos~\etal~\cite{pavlakos17volumetric} might lead to slightly better results on EMDB.
The distribution accuracy metrics for 3DPW~\cite{marcard2018eccv} are provided by ScoreHypo and we outperform them by a large margin.

To generate the qualitative results for ProHMR~\cite{kolotouros2021prohmr} in Fig.~1 and Fig.~2 of the main paper, we use our retrained baseline model ProHMR\textsuperscript{\textdagger}. 
This baseline is trained on the same three datasets using the same image backbone as our proposed model, and is more accurate than the officially released checkpoint.

\section{Additional Quantitative Results}
\begin{figure}
\centering		
\begin{tikzpicture}[scale=0.8]
		\begin{axis}[ 
		legend entries={HuManiFlow*~\cite{sengupta2023humaniflow},
		    ProHMR\textsuperscript{\textdagger}~\cite{kolotouros2021prohmr},
		    Ours,},
		legend cell align={left},
		xmax=1000,
		xmin=0,
		ymax=85,
		ymin=45,
		legend pos=north east,
		samples=100,
		xtick={0,200,...,1000},
		ytick={45,50,...,85},
		grid=major,
        title=3DPW,
		xlabel={number of hypotheses},
		y label style={at={(axis description cs:0.09,.5)},anchor=south},
		ylabel={PVE ($mm$)}
		] 
		\addplot[color=green, style=thick] table [x=n_hypos, y=humaniflow, col sep=comma] {plots/error_list_plot.csv};
		\addplot[color=red, style=thick] table [x=n_hypos, y=prohmr, col sep=comma] {plots/error_list_plot.csv};
	\addplot[color=blue, style=thick] table [x=n_hypos, y=ours, col sep=comma] {plots/error_list_plot.csv};

	\end{axis}
		\end{tikzpicture}
\caption{Evaluation results on 3DPW for an increasing number of generated 3D human mesh hypotheses.
}
\label{fig:hypoVsError}
\end{figure}
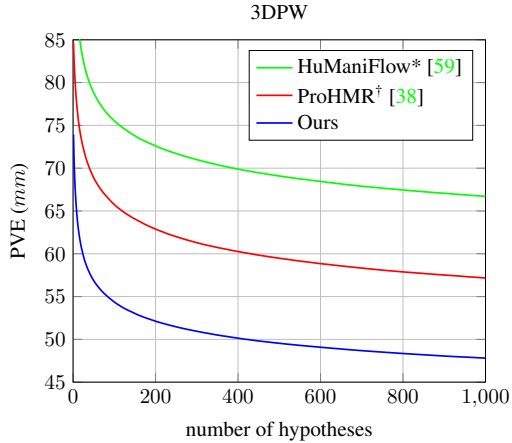

\paragraph{Number of hypotheses.} 
Fig.~\ref{fig:hypoVsError} shows the Per Vertex Error (PVE) for an increasing amount of hypotheses on 3DPW.
The PVE continues to improve significantly when generating more than 100 hypotheses, reaching a PVE of $47.8\,mm$ for 1000 samples compared to $54.4\,mm$ for 100.

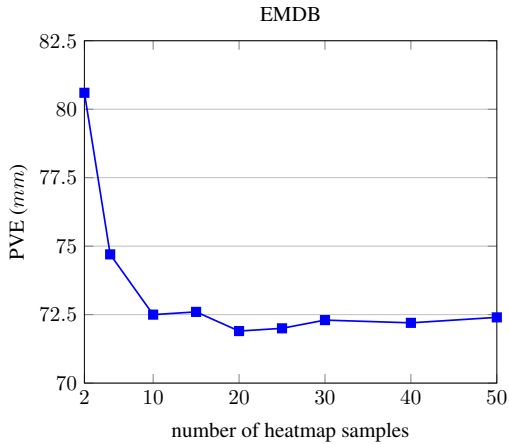
\begin{figure}
\centering		
\begin{tikzpicture}[scale=0.8]
\begin{axis}[
    title={EMDB},
    xlabel={number of heatmap samples},
    ylabel={PVE ($mm$)},
    y label style={at={(axis description cs:0.06,.5)},anchor=south},
    samples=100,
    xmin=2, xmax=50,
    ymin=70, ymax=82.5,
    xtick={2,10,20,30,40,50},
    ytick={70, 72.5, 75, 77.5, 80, 82.5, 85},
    ymajorgrids=true,
]

\addplot[
    color=blue,
    mark=square*,
    style=thick,
    ]
    coordinates {
    (2,80.6)(5,74.7)(10,72.5)(15,72.6)(20,71.9)(25,72.0)(30,72.3)(40,72.2)(50,72.4)
    };
    
\end{axis}
\end{tikzpicture}
\caption{Evaluation results on EMDB for an increasing number of joint samples drawn from the heatmaps for calculating $\mathcal{L}_{\text{MMD}}$.
The minimum Per Vertex Error (PVE) of 100 hypotheses is evaluated.
Each square denotes a model trained with the specified number of heatmap samples.
}
\label{fig:num_hm_samples}
\end{figure}

\paragraph{Number of heatmap samples.}
We utilize heatmaps of the 2D pose detector ViTPose~\cite{xu22vitpose} to directly supervise the learned distributions of our model using the sample-based loss $\mathcal{L}_{\text{MMD}}$. 
The loss computes the Maximum Mean Discrepancy between samples drawn from heatmaps and 2D reprojections of random NF hypotheses.
To analyze the influence of the number of samples used for $\mathcal{L}_{\text{MMD}}$, we show the performance for different configurations in Fig.~\ref{fig:num_hm_samples}. 
The performance first improves with an increasing number of samples, and then remains stable over a wide range.
When using only very few samples for computing $\mathcal{L}_{\text{MMD}}$, the model cannot successfully learn to reproduce the distributions encoded in the heatmaps and often predicts distributions with very low diversity.
Intuitively, a sufficient number of samples is required to represent the heatmap distributions, while the computational complexity grows with increasing number of samples.
As a good trade-off, we use 25 samples in all other experiments.

\paragraph{Ablation study on EMDB.}
We conduct the ablation study of the main paper on EMDB and present the results in Table~\ref{table:ablations_suppl}.
Our proposed design choices and loss functions all contribute to the accuracy of the predicted distributions.
Notably, despite being added last in the ablation study, the use of $\mathcal{L}_{\text{MMD}}$ results in large improvements.

\begin{table}
	\centering
 \resizebox{0.9\linewidth}{!}{
	\begin{tabular}{lccccc}
		\toprule
       & & EMDB (24) \\
		\cmidrule(lr){2-6}
		  Models &
		\multicolumn{1}{c}{MPJPE $\downarrow$} & \multicolumn{1}{c}{PA-MPJPE $\downarrow$} & \multicolumn{1}{c}{PVE $\downarrow$} & \multicolumn{1}{c}{} & \multicolumn{1}{c}{}  \\
		\midrule
        ProHMR\textsuperscript{\textdagger}~\cite{kolotouros2021prohmr} & 76.7 & 47.1 & 87.3\\
        + bbox info~\cite{li2022cliff} & 73.1 & 46.5 & 82.3 \\
        + 2D pose condition & 69.0 & 44.0 & 77.9 \\
        + RealNVP & 68.5 & 43.1 & 77.5 \\
        + $\mathcal{L}_{\text{MMD}}$ & 63.9 & 40.7 & 72.4 \\
        + $\mathcal{L}_{\text{mask}}$ (Ours full) & 63.6 & 40.9 & 72.0\\
		\bottomrule
	\end{tabular}
    }
 	\caption{Ablation study analyzing our proposed design choices and loss functions.
  Components are added successively, and the minimum errors out of 100 hypotheses are reported.
  }
  	\label{table:ablations_suppl}
\end{table}

\begin{table*}
	\centering
 \resizebox{0.75\linewidth}{!}{
	\begin{tabular}{lccc|ccc}
		\toprule
       Supervising & & 3DPW (14) & & & EMDB (24) \\
		\cmidrule(lr){2-4}\cmidrule(lr){5-7}
		   random hypotheses &
		\multicolumn{1}{c}{MPJPE $\downarrow$} & \multicolumn{1}{c}{PA-MPJPE $\downarrow$} & \multicolumn{1}{c}{PVE $\downarrow$} & \multicolumn{1}{c}{MPJPE $\downarrow$} & \multicolumn{1}{c}{PA-MPJPE $\downarrow$} & \multicolumn{1}{c}{PVE $\downarrow$}  \\
		\midrule
        no supervision & 48.9 & 32.1 & 57.4 & 67.8 & 43.2 & 76.8 \\
        \hdashline
        $\mathcal{L}_{\text{2D-all}}$, $\lambda = 1e^{-3}$ & 48.1 & 31.8 & 56.9 & 68.5 & 43.1 & 77.5 \\
        $\mathcal{L}_{\text{2D-all}}$, $\lambda = 5e^{-3}$ & 51.3 & 33.3 & 60.5 & 71.6 & 45.4 & 81.0 \\
        $\mathcal{L}_{\text{2D-all}}$, $\lambda = 1e^{-2}$ & 53.2 & 34.2 & 62.9 & 71.3 & 46.2 & 80.8 \\
        \hdashline
        $\mathcal{L}_{\text{2D-vis}}$, $\lambda = 1e^{-3}$ & 48.4 & 31.9 & 57.3 & 68.7 & 43.0 & 77.4 \\
        $\mathcal{L}_{\text{2D-vis}}$, $\lambda = 5e^{-3}$ & 47.6 & 31.4 & 56.3 & 67.9 & 42.3 & 76.6 \\
        $\mathcal{L}_{\text{2D-vis}}$, $\lambda = 1e^{-2}$ & 50.6 & 32.7 & 59.5 & 71.3 & 45.4 & 80.8 \\
        \hdashline
        DiffPose condition~\cite{holmquist2023diffpose} & 48.1 & 32.1 & 57.0 & 68.7 & 42.9 & 77.4\\
        \hdashline
        $\mathcal{L}_{\text{MMD}}$ (Ours) & $\bm{46.5}$ & $\bm{29.7}$ & $\bm{54.8}$ & $\bm{63.9}$ & $\bm{40.7}$ & $\bm{72.4}$ \\
		\bottomrule
	\end{tabular}
}
 	\caption{Evaluation results for the ablation study on how to best supervise random hypotheses during training.
  The minimum errors out of 100 hypotheses are reported.
  Random samples are either not supervised, supervised by minimizing an $l_1$ loss to the ground-truth 2D positions for either all ($\mathcal{L}_{\text{2D-all}}$) or only visible ($\mathcal{L}_{\text{2D-vis}}$) 2D joints, or by using our proposed $\mathcal{L}_{\text{MMD}}$ loss.} 
  	\label{table:evalmmd}
\end{table*}

\paragraph{Detailed $\mathcal{L}_{\text{MMD}}$ ablation study.}
A main contribution of this work is to directly supervise the learned distributions by minimizing the distance to distributions encoded in heatmaps of a 2D pose detector~\cite{xu22vitpose} using the sample-based distance measure $\mathcal{L}_{\text{MMD}}$.
To further analyze the influence of $\mathcal{L}_{\text{MMD}}$, we perform additional experiments on 3DPW and EMDB.
The goal is to evaluate different ways of supervising random hypotheses generated by the normalizing flow during training.
The mask loss $\mathcal{L}_{\text{mask}}$ is not applied in this study.
As a baseline, we first train a model without $\mathcal{L}_{\text{MMD}}$ and where the 2D reprojection loss $\mathcal{L}_{\text{2D}}$ is only computed for the approximated mode prediction.
Random NF hypotheses are thus not supervised for this model.
Based on this baseline, we train models where either all joints ($\mathcal{L}_{\text{2D-all}}$) or only visible joints ($\mathcal{L}_{\text{2D-vis}}$) of random hypotheses are penalized by minimizing the distance to the ground-truth 2D joints using an $l_1$ loss.
This is done by ProHMR~\cite{kolotouros2021prohmr} and HuManiFlow~\cite{sengupta2023humaniflow}, respectively.
Furthermore, we train a model that receives the embedding proposed in DiffPose~\cite{holmquist2023diffpose} as additional condition, which is computed based on samples drawn from the heatmaps.
2D reprojections of random hypotheses are not penalized during training of this model.
The distribution accuracy metrics for 100 hypotheses per image are presented in Table~\ref{table:evalmmd}.
Supervising all 2D joints of random hypotheses by minimizing the distance to the ground-truth position has overall no positive impact on the distribution accuracy.
On the contrary, it heavily restricts the learned distributions, leading to low sample diversity.
When only supervising visible joints using a loss weight of $\lambda = 5e^{-3}$, the metrics slightly improve.
While it is intuitive to enforce all visible joints to be at the 2D location of the ground-truth, we find that this also leads to significantly less diversity generated for invisible joints, which has negative influence on the distribution accuracy.
Moreover, performance of the models heavily depends on the 2D loss weight.
Using DiffPose embeddings as additional condition does not lead to improvements in our setting.
Note that in contrast to our setting, the original DiffPose model does not use image features as condition and thus has more incentives to process and exploit the information encoded in the embeddings.
Finally, joint-wise minimizing the Maximum Mean Discrepancy between 2D reprojections of random hypotheses and samples drawn from heatmaps consistently leads to learned distributions with the highest accuracy. 
With $\mathcal{L}_{\text{MMD}}$, the learned distributions are explicitly optimized to have high diversity for ambiguous and low diversity for unambiguous joints.

\section{Additional Qualitative Results}

In the following, we will present additional qualitative results.
Predicted camera parameters are used for rendering the 3D human mesh hypotheses and a side-view of each human mesh is created by a rotation of $90^{\circ}$ or $270^{\circ}$ around the y-axis in camera space.

\begin{figure*}
\centering
\includegraphics[width=1.0\linewidth]{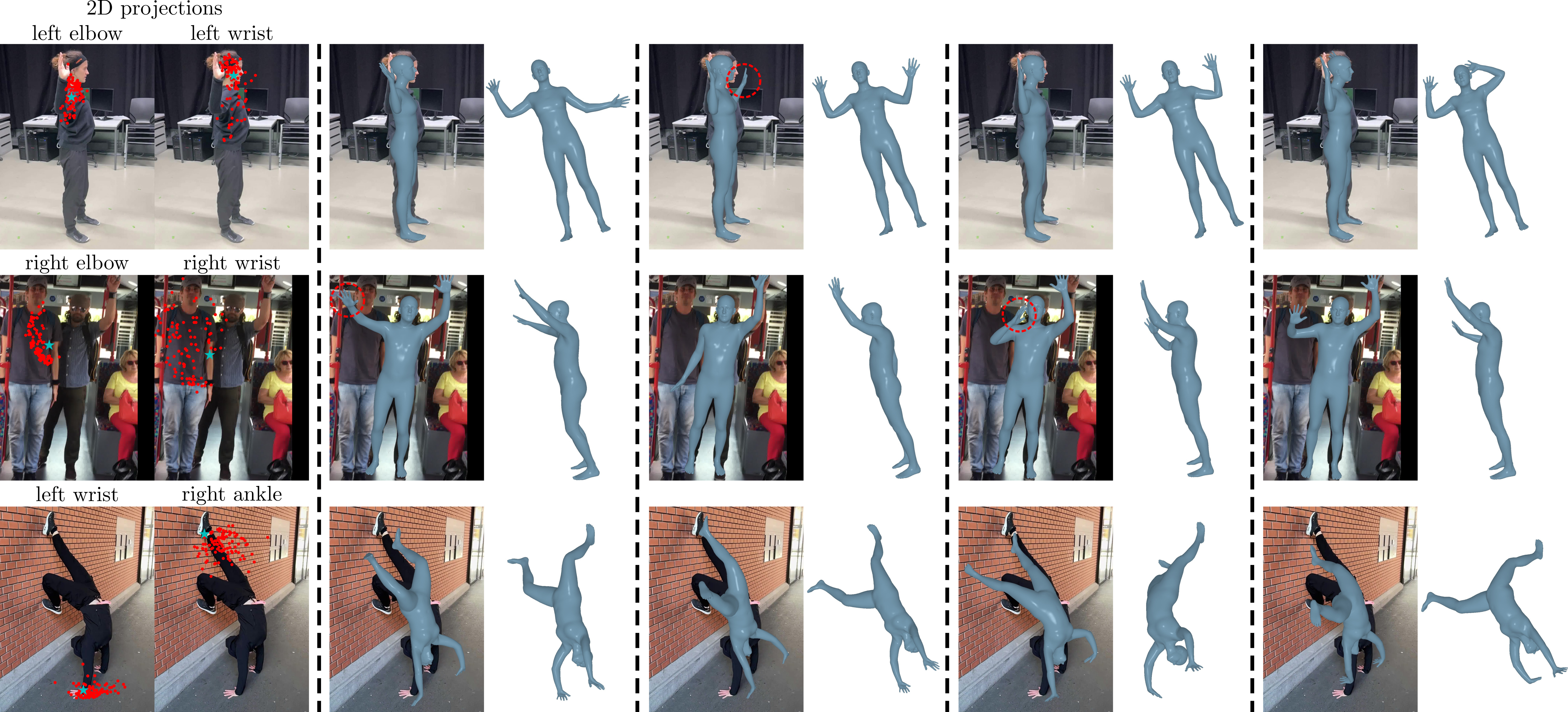}
\caption{Typical failure cases of our approach.
For highly ambiguous joints, our model predicts distributions with very high diversity, sometimes containing a few incorrect samples highlighted with a red circle (rows 1 and 2).
The model fails to predict meaningful distributions for very unusual poses not seen during training (row 3).
}
\label{fig:failure_cases}
\end{figure*}

\paragraph{Failure cases.}
A few examples of undesirable behavior of our model are depicted in Fig.~\ref{fig:failure_cases}.
While optimizing $\mathcal{L}_{\text{mask}}$ significantly decreases the number of incorrect hypotheses, the model still sometimes generates hypotheses where joints are visible that should be invisible.
This typically happens for highly ambiguous joints for which the model predicts distributions with very high diversity. 
We find that using a larger loss weight $\lambda_{\text{mask}}$ for the mask loss can further decrease the number of incorrect hypotheses.
However, this comes at a cost of reducing the diversity of the learned distributions too much, resulting in worse accuracy metrics.
Finding a way to further reduce the number of incorrect hypotheses while maintaining meaningful diversity could be promising future work.
Another typical failure case occurs when the model is presented with highly unusual poses not seen during training. 
For such examples, high diversity is generated even for unambiguous joints.
However, in contrast to deterministic regressors, our model provides information about the prediction uncertainty, either by computing the variance of the hypotheses or by directly calculating their likelihoods.
This is useful for downstream tasks that need to know whether the reconstructions results are accurate or not.

\begin{figure*}
\centering
\includegraphics[width=1.0\linewidth]{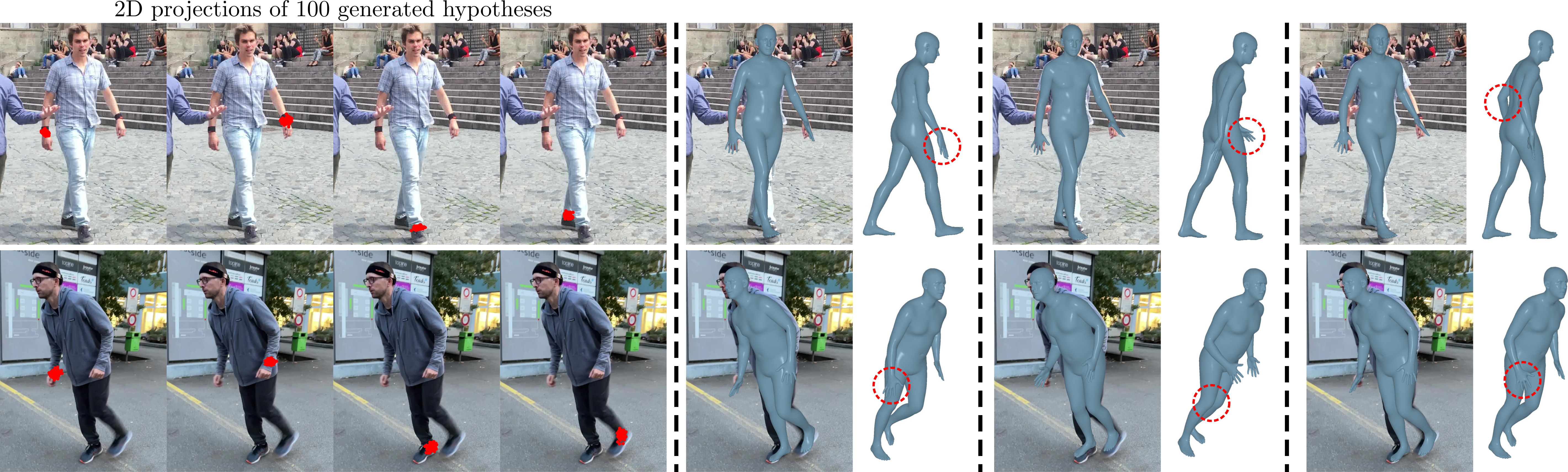}
\caption{Examples demonstrating depth ambiguity for monocular 3D human mesh estimation.
Although all hypotheses vary only slightly along the image directions, significant diversity for the depth is generated.
Reprojections of 100 hypotheses for the right wrist, left wrist, right ankle, and left ankle are shown.}
\label{fig:depth_ambi}
\end{figure*}

\paragraph{Depth ambiguity.}
Even if all body parts of the person are clearly visible in the image, the depth often cannot be uniquely reconstructed.
We show two of such examples in Fig.~\ref{fig:depth_ambi}.
The predicted hypotheses vary only slightly along the image directions, but have high variance for the depth.

\begin{figure*}
\centering
\includegraphics[width=1.0\linewidth]{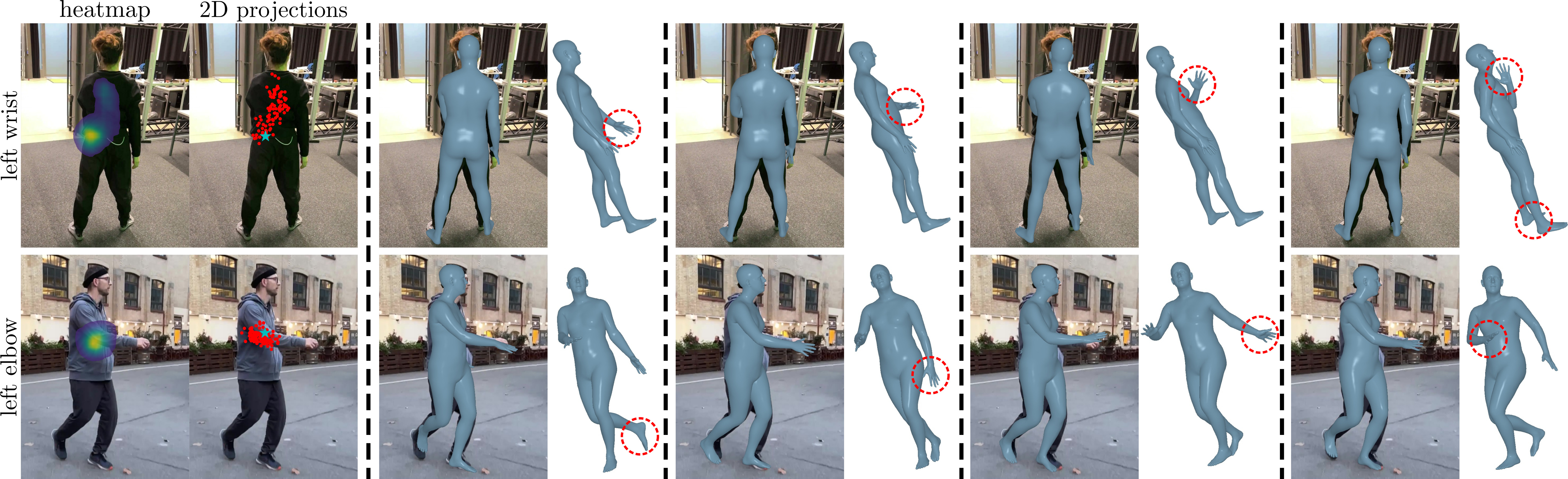}
\caption{Predicted heatmaps of ViTPose~\cite{xu22vitpose} are shown together with 3D human mesh hypotheses generated by our model.}
\label{fig:hm_variance}
\end{figure*}

\paragraph{Uncertainty in heatmaps of ViTPose~\cite{xu22vitpose}.}
We visualize heatmap predictions of ViTPose for occluded joints together with 3D mesh hypotheses generated by our model in Fig.~\ref{fig:hm_variance}. 
The predicted heatmaps encode meaningful joint uncertainty information that our model successfully utilizes during training.

\begin{figure*}
\centering
\includegraphics[width=1.0\linewidth]{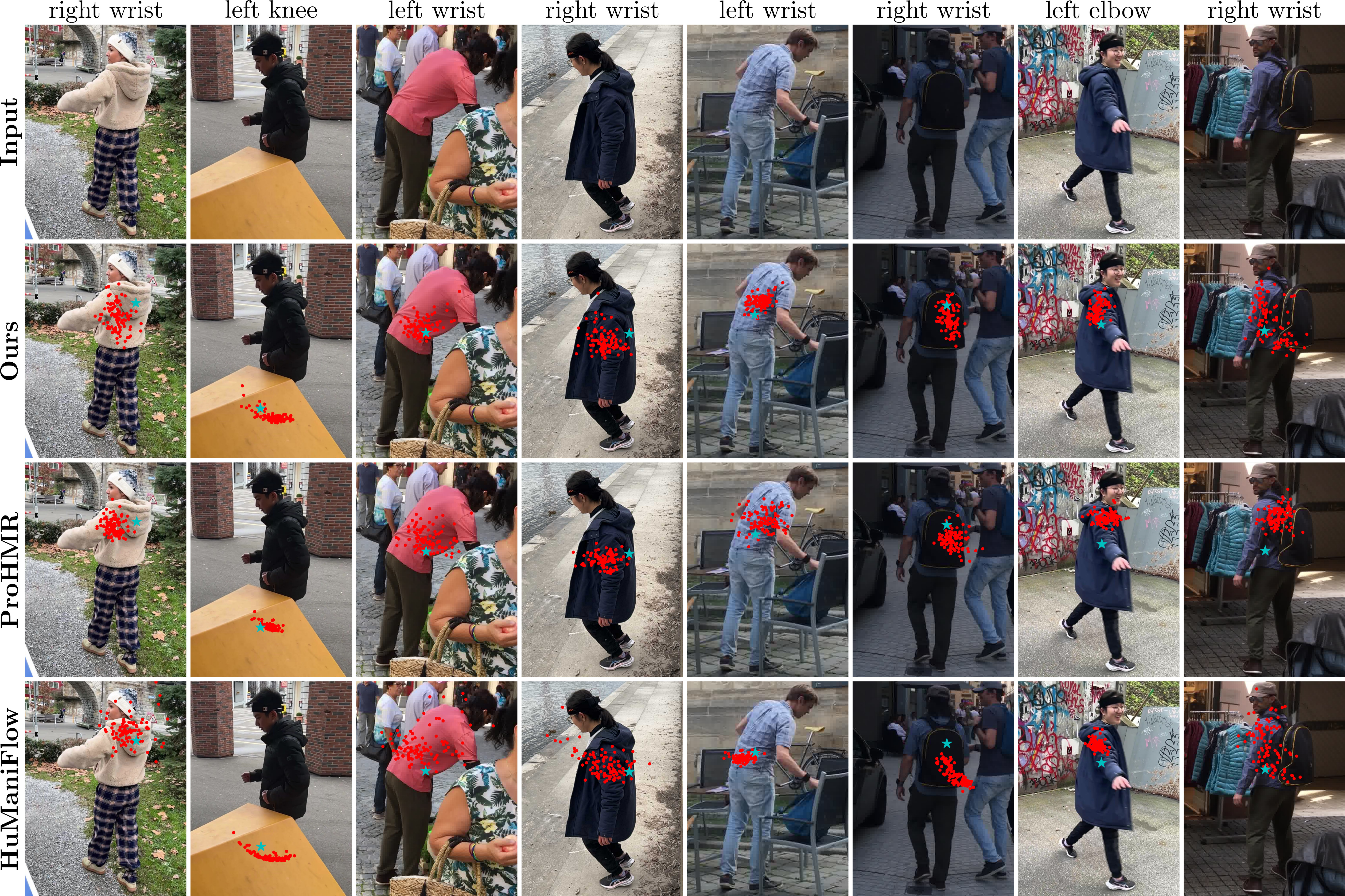}
\caption{Qualitative comparison with the competing methods ProHMR~\cite{kolotouros2021prohmr} and HuManiFlow~\cite{sengupta2023humaniflow}.
The 2D reprojections of 100 hypotheses for highly ambiguous joints are shown.}
\label{fig:competitors}
\end{figure*}

\paragraph{Comparison with competitors.}
We qualitatively compare the performance of our model with ProHMR\textsuperscript{\textdagger} and HuManiFlow by visualizing the reprojections of 100 hypotheses for highly ambiguous joints in Fig.~\ref{fig:competitors}.
Our model generates more plausible and more meaningfully diverse 3D human mesh hypotheses than the competitors.

\section{Limitations and Future Work}
Following previous work~\cite{sengupta2023humaniflow}, we define a joint to be invisible if the corresponding heatmap predicted by a 2D pose detector has a maximum value below a certain threshold.
While this works well for most cases, we observe that the 2D detector sometimes tends to be overconfident. 
This calibration gap in 2D human pose estimation frameworks was also recently observed and analyzed by Gu~\etal~\cite{gu24calibration}.
Future work could examine using explicitly predicted joint visibility scores~\cite{gu24calibration,sun2024visibility} instead of maximum heatmap values to decide if a joint is invisible for training and evaluation. 

Since we use the distributions encoded in the heatmaps of a 2D pose detector as supervision signal, the performance of our model is influenced by the accuracy of these encoded distributions.
Thus, another interesting future research direction would be to improve the distribution modeling capabilities of 2D pose estimators.

\end{document}